\documentclass[runningheads]{llncs}
\usepackage{graphicx}
\usepackage{amsmath,graphicx}
\usepackage{times}
\usepackage{latexsym}
\usepackage{times}
\usepackage{latexsym}
\usepackage{subfigure}
\usepackage{caption}
\usepackage{comment}
\usepackage{times}
\usepackage{soul}
\usepackage{booktabs}
\usepackage{url}
\usepackage[utf8]{inputenc}
\usepackage{graphicx}
\usepackage{amsmath}
\usepackage{amssymb}
\usepackage{booktabs}
\usepackage{algorithm}
\usepackage{algorithmic}
\usepackage{multirow}
\usepackage{multicol}
\usepackage{caption}
\usepackage{xspace}

\begin{document}
\title{A Position Aware Decay Weighted Network \\ for Aspect based Sentiment Analysis}%

\author{
Avinash Madasu \and Vijjini Anvesh Rao}
\institute{Samsung R\&D Institute, Bangalore\\
           \email{\{m.avinash,a.vijjini\}@samsung.com}}
\maketitle              
\begin{abstract}
Aspect Based Sentiment Analysis (ABSA) is the task of identifying sentiment polarity of a text given another text segment or aspect. In ABSA, a text can have multiple sentiments depending upon each aspect. Aspect Term Sentiment Analysis (ATSA) is a subtask of ABSA, in which aspect terms are contained within the given sentence. Most of the existing approaches proposed for ATSA, incorporate aspect information through a different subnetwork thereby overlooking the advantage of aspect terms' presence within the sentence. In this paper, we propose a model that leverages the positional information of the aspect. The proposed model introduces a decay mechanism based on position. This decay function mandates the contribution of input words for ABSA. The contribution of a word declines as farther it is positioned from the aspect terms in the sentence. The performance is measured on two standard datasets from SemEval 2014 Task 4. In comparison with recent architectures, the effectiveness of the proposed model is demonstrated.

\end{abstract}
\section{Introduction}
\label{sec:intro}
Text Classification deals with the branch of Natural Language Processing (NLP) that involves classifying a text snippet into two or more predefined categories. Sentiment Analysis (SA) addresses the problem of text classification in the setting where these predefined categories are sentiments like positive or negative \cite{pang2002thumbs}. 
Aspect Based Sentiment Analysis (ABSA) is proposed to perform sentiment analysis at an aspect level \cite{hu2004mining}. 
There are four sub-tasks in ABSA namely Aspect Term Extraction (ATE), Aspect Term Sentiment Analysis (ATSA),  Aspect Category Detection (ACD), Aspect Category Sentiment Analysis (ACSA). In the first sub-task (ATE), the goal is to identify all the aspect terms for a given sentence. 
Aspect Term Sentiment Analysis (ATSA) is a classification problem where given an aspect and a sentence, the sentiment has to classified into one of the predefined polarities. In the ATSA task, the aspect is present within the sentence but can be a single word or a phrase. In this paper, we address the problem of ATSA.
Given a set of aspect categories and a set of sentences, the problem of ACD is to classify the aspect into one of those categories. ACSA can be considered similar to ATSA, but the aspect term may not be present in the sentence.  
It is much harder to find sentiments at an aspect level compared to the overall sentence level because the same sentence might have different sentiment polarities for different aspects. For example consider the sentence, \textit{"The taste of food is good but the service is poor"}. If the aspect term is \textit{food}, the sentiment will be \textit{positive}, whereas if the aspect term is \textit{service}, sentiment will be \textit{negative}. Therefore, the crucial challenge of ATSA is modelling the relationship between aspect terms and its context in the sentence. Traditional methods involve feature engineering trained with machine learning classifiers like Support Vector Machines (SVM) \cite{kiritchenko2014nrc}. However, these methods do not take into account the sequential information and require a considerable struggle to define the best set of features. With the advent of deep learning, neural networks are being used for the task of ABSA.
For ATSA, LSTM coupled with attention mechanism \cite{bahdanau2014neural} have been widely used to focus on words relevant to certain aspect. Target-Dependent Long Short-Term Memory (TD-LSTM) uses two LSTM networks to model left and right context words surrounding the aspect term \cite{tang2015effective}. The outputs from last hidden states of LSTM are concatenated to find the sentiment polarity. Attention Based LSTM (ATAE-LSTM) uses attention on the top of LSTM to concentrate on different parts of a sentence when different aspects are taken as input \cite{wang2016attention}. Aspect Fusion LSTM (AF-LSTM) \cite{tay2018learning} uses associative relationship between words and aspect to perform ATSA. Gated Convolution Neural Network (GCAE) \cite{xue2018aspect} employs a gated mechanism to learn aspect information and to incorporate it into sentence representations.

However, these models do not utilize the advantage of the presence of aspect term in the sentence. They either employ an attention mechanism with complex architecture to learn relevant information or train two different architectures for learning sentence and aspect representations. In this paper, we propose a model that utilizes the positional information of the aspect in the sentence. We propose a parameter-less decay function based learning that leverages the importance of words closer to the aspect. Hence, evading the need for a separate architecture for integrating aspect information into the sentence. The proposed model is relatively simple and achieves improved performance compared to models that do not use position information. We experiment with the proposed model on two datasets, restaurant and laptop from SemEval 2014.

\section{Related Work}
\subsection{Aspect Term Sentiment Analysis} 
Early works of ATSA, employ lexicon based feature selection techniques like Parts of Speech Tagging (POS), unigram features and bigram features \cite{kiritchenko2014nrc}. However, these methods do not consider aspect terms and perform sentiment analysis on the given sentence. \newline Phrase Recursive Neural Network for Aspect based Sentiment Analysis (PhraseRNN) \cite{nguyen-shirai-2015-phrasernn} was proposed based on Recursive Neural Tensor Network \cite{socher-etal-2013-recursive} primarily used for semantic compositionality. PhraseRNN uses dependency and constituency parse trees to obtain aspect representation. An end-to-end neural network model was introduced for jointly identifying aspect and polarity \cite{schmitt-etal-2018-joint}. This model is trained to jointly optimize the loss of aspect and the polarity. In the final layer, the model outputs one of the sentiment polarities along with the aspect. \cite{AAAI1816570} introduced Aspect Fusion LSTM (AF-LSTM) for performing ATSA.

\section{Model}
In this section, we propose the model Position Based Decay Weighted Network (PDN). The model architecture is shown in Figure \ref{fig:architecture}. The input to the model is a sentence $S$ and an Aspect $A$ contained within it. Let $n$ represent the maximum sentence length considered.
\subsection{Word Representation}
Let V be the vocabulary size considered and $X$ $\in$ $\mathbb{R}^{V \times d_{w}}$ represent the embedding matrix\footnote{https://nlp.stanford.edu/data/glove.840B.300d.zip}, where for each word $X_{i}$ is a $d_{w}$ dimensional word vector. Words contained in the embedding matrix are initialized to their corresponding vectors whereas words not contained are initialized to 0's. $I$ $\in$ $\mathbb{R}^{n \times d_{w}}$ denotes the pretrained embedding representation of a sentence where $n$ is the maximum sentence length. 
\subsection{Position Encoding}
In the ATSA task, aspect $A$ is contained in the sentence $S$. A can be a word or a phrase. Let $k_{s}$ denote the starting index and $k_{e}$ denote the ending index of the aspect term(s) in the sentence. Let $i$ be the index of a word in the sentence. The position encoding of words with respect to aspect are represented using the formula 
\begin{equation}
  p(i) =\left\{
  \begin{array}{@{}ll@{}}
    k_{s}-i+1, & \ k_{s}>i \\
    1, & \ i \in k_{s},k_{s+1},..,k_{e-1},k_{e} \\
    i-k_{e}+1, & \ i>k_{e} 
  \end{array}\right.
\end{equation} 
The position encodings for the sentence ``granted the space is smaller than most it is the best service" where ``space" is the aspect is shown in Figure \ref{fig:architecture}. This number reflects the relative distance of a word from the closest aspect word. The position embeddings from the position encodings are randomly initialized and updated during training. Hence, $P$ $\in$ $\mathbb{R}^{n \times d_{p}}$ is the position embedding representations of the sentence. $d_{p}$ denotes the number of dimensions in the position embedding.
\subsection{Architecture}
As shown in Figure \ref{fig:architecture}, PDN comprises of two sub-networks: Position Aware Attention Network(PDN) and Decay Weighting Network (DWN).
\subsubsection*{Position Aware Attention Network (PAN)}
An LSTM layer is trained on $I$ to produce hidden state representation $h_{t}$ $\in$ $\mathbb{R}^{d_{h}}$ for each time step $t$ $\in$ $\{1,n\}$ where $d_{h}$ is the number of units in the LSTM. The LSTM outputs contain sentence level information and Position embedding contain aspect level information. An attention subnetwork is applied on all $h$ and $P$ to get a scalar score $\alpha$ indicating sentiment weightage of the particular time step to the overall sentiment. However, prior to concatenation, the position embeddings and the LSTM outputs may have been output from disparate activations leading to different distribution. Training on such values may bias the network towards one of the representations. Therefore, we apply a fully connected layer separately but with same activation function Scaled Exponential Linear Unit (SELU)\cite{klambauer2017self} upon them. Two fully connected layers follow this representation. Following are the equations that produce $\alpha$ from LSTM outputs $h$ and position embeddings $P$.  
\begin{equation}
    P_{t}^{\prime} = selu(W_{p} \cdot P_{t} + b_{p})
\end{equation}
\begin{equation}
    h_{t}^{\prime} = selu(W_{h} \cdot h_{t} + b_{h})
\end{equation}
\begin{equation}
    H_{t} = relu(W_{a} \cdot [h_{t}^{\prime}P_{t}^{\prime}] + b_{a})
\end{equation}
\begin{equation}
    e_{t} = \tanh( \mathbf{v_{}}^\intercal \cdot H_{t})
\end{equation}
\begin{equation}
    \alpha_{t} = \frac{\exp(e_{t})}{\sum_{i=1}^{n}\exp(e_{i})}
\end{equation}
\subsubsection*{Decay Weighting Network (DWN)}
In current and following section, we introduce decay functions. The decay function for scalar position encoding $p(i)$ is represented as the scalar $d(p(i))$. These functions are continuously decreasing in the range $[0,\infty)$. The outputs from the LSTM at every time step are scaled by the decay function's output.
\begin{equation}
    Z_{t} = h_{t} \cdot d(p(t)) \: \forall  \: t \in \{1,n\}
\end{equation}
A weighted sum $O$ is calculated on the outputs of Decay Weighted network using the attention weights from PAN.
\begin{equation}
    O = \alpha \cdot Z
\end{equation}
A fully connected layer is applied on $O$ which provides an intermediate representation $Q$. A softmax layer is fully connected to this layer to provide final probabilities.

\begin{figure*}
\centering
\includegraphics[width=4cm]{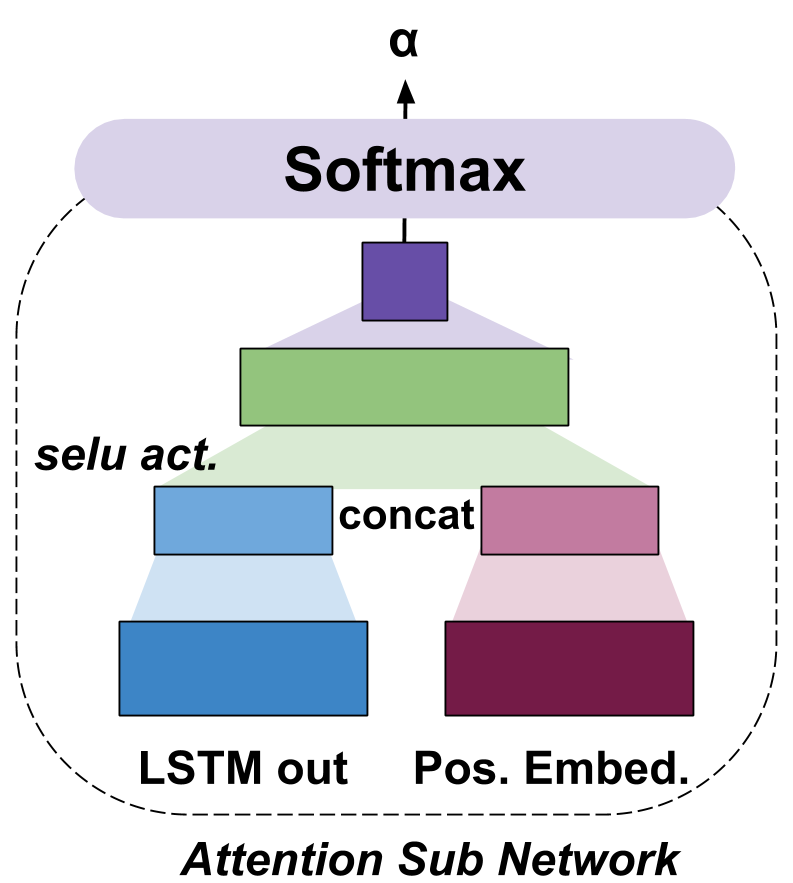}
\caption{Attention Sub Network}
\label{fig:subnetwork}
\end{figure*}

It is paramount to note that the DWN does not contain any parameters and only uses a decay function and multiplication operations. The decay function provides us with a facility to automatically weight representations closer to aspect as higher and far away as lower, as long as the function hyperparameter is tuned fittingly. Lesser parameters makes the network efficient and easy to train.

\begin{figure*}
\centering
\includegraphics[width=12cm]{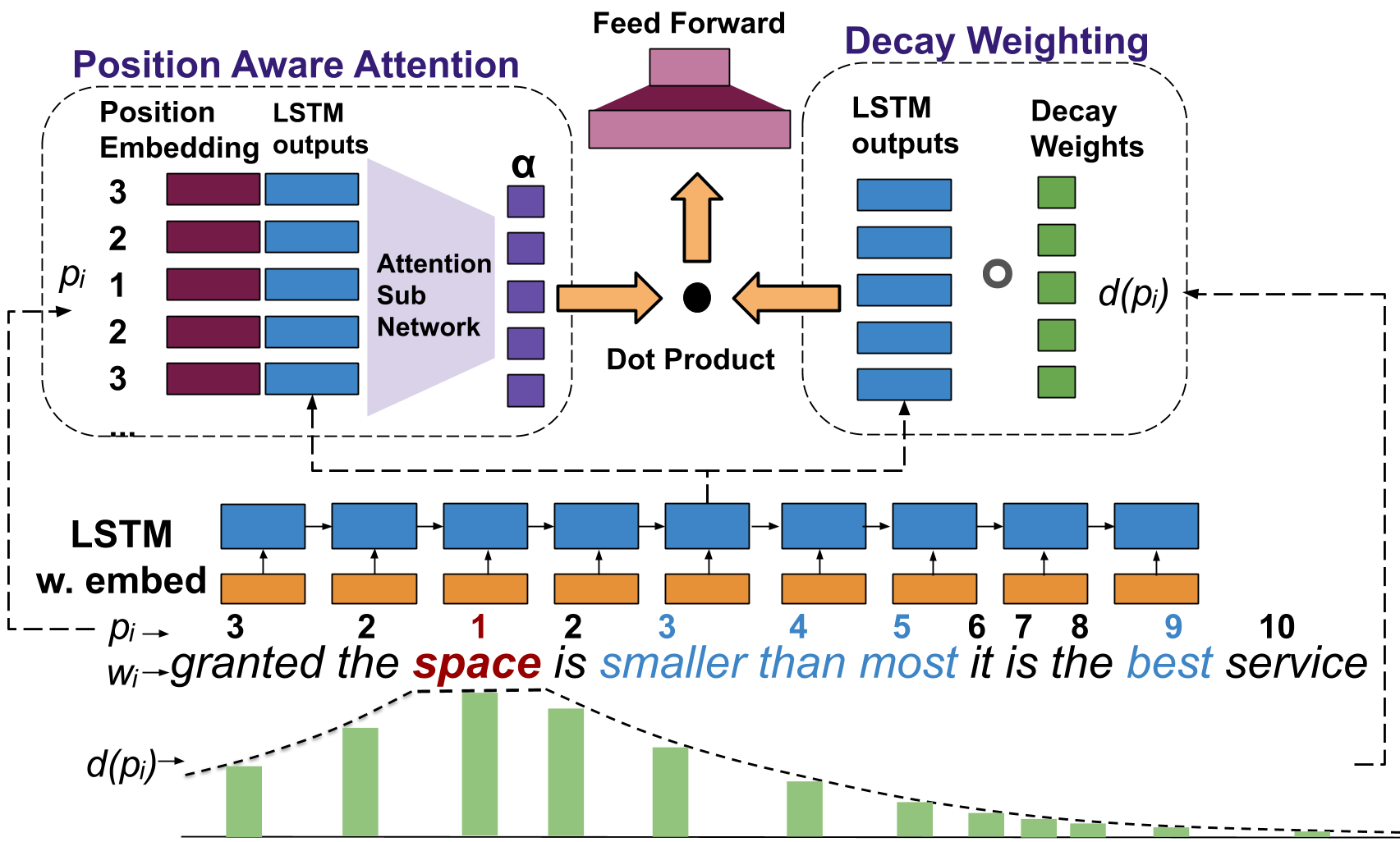}
\caption{PDN Architecture, in the shown example, ``space" is the aspect. Refer to Figure \ref{fig:subnetwork} for the Attention Sub Network.}
\label{fig:architecture}
\end{figure*}

\begin{table}[t!]
\begin{center}
\begin{tabular}{l|c|c}
\hline \bf Model & \bf Restaurant & \bf Laptop \\ \hline
Majority &65.00 &53.45 \\
NBOW &67.49 &58.62 \\
LSTM & 67.94& 61.75\\
TD-LSTM  & 69.73 &62.38\\
AT-LSTM & 74.37 & 65.83 \\
ATAE-LSTM  & 70.71 & 60.34 \\
DE-CNN & 75.18 & 64.67 \\
AF-LSTM & 75.44 & 68.81 \\
GCAE & 76.07 & 67.27 \\
\hline
Tangent-PDN & 78.12 & 68.82\\
Inverse-PDN & \textbf{78.9} & \textbf{70.69} \\
Expo-PDN & 78.48 & 69.43 \\
\hline
\end{tabular}
\end{center}
\caption{\label{tab:acc} Accuracy Scores of all models. Performances of
baselines are cited from \cite{tay2018learning} }
\end{table}
\subsubsection*{Decay Functions}
We performed experiments with the following decay functions. \\
{\bf Inverse Decay}:\\
Inverse decay is represented as:
\begin{equation}
    d(x) = \frac{\lambda}{x}
\end{equation}
{\bf Exponential Decay}: \\
Exponential decay is represented as:
\begin{equation}
    d(x) = e^{-\lambda * x}
\end{equation} 
{\bf Tangent Decay}: \\
Tangent decay is represented as:
\begin{equation}
    d(x) = 1-tanh(\lambda*x)
\end{equation}
$\lambda$ is the hyper-parameter in all the cases.\footnote{In our experiments we took $\lambda$ = 0.45 for Tangent-PDN, 1.1333 for Inverse-PDN and 0.3 for Expo-PDN }

\section{Experiments}
\subsection{Datasets}
We performed experiments on two datasets, Restaurant and Laptop from SemEval 2014 Task 4 \cite{pontiki-etal-2014-semeval}. Each data point is a triplet of sentence, aspect and sentiment label. The statistics of the datasets are shown in the Table \ref{statistics}. As most existing works reported results on three sentiment labels $\textit{positive,negative,neutral}$ we performed experiments by removing \textit{conflict} label as well.
\subsection{Compared Methods}
We compare proposed model to the following baselines:
\subsubsection{Neural Bag-of-Words (NBOW)}  NBOW  is the sum of word embeddings in the sentence \cite{tay2018learning}. 
\subsubsection{LSTM}  Long Short Term Memory (LSTM) is an important baseline in NLP. For this baseline, aspect information is not used and sentiment analysis is performed on the sentence alone. \cite{tay2018learning}. 
\subsubsection{TD-LSTM}  In TD-LSTM, two separate LSTM layers for modelling the preceding and following contexts of the aspect is done for aspect sentiment analysis \cite{tang2015effective}. 
\subsubsection{AT-LSTM} In Attention based LSTM (AT-LSTM), aspect embedding is used as the context for attention layer, applied on the sentence \cite{wang2016attention}. 
\subsubsection{ATAE-LSTM} 
In this model, aspect embedding is concatenated with input sentence embedding. LSTM is applied on the top of concatenated input \cite{wang2016attention}. 

\begin{table*}
\begin{center}
  \begin{tabular}{l|l|l|l|l|l|l}
    \hline
    \multirow{2}{*}{Dataset} &
    \multicolumn{2}{c|}{Positive} & \multicolumn{2}{c|}{Negative} & \multicolumn{2}{c}{Neutral} \\
    \cline{2-7}
    & Train & Test & Train & Test & Train & Test \\
    \hline
    Restaurant &2164  & 728 & 805& 196 & 633&  196\\
    \hline
    Laptop & 987  & 341  & 866  & 128  & 460  & 169  \\
    \hline
  \end{tabular}
  \end{center}
  \caption{\label{statistics} Statistics of the datasets}
\end{table*}

\subsubsection{DE-CNN} 
Double Embeddings Convolution Neural Network (DE-CNN) achieved state of the art results on aspect extraction. We compare proposed model with DE-CNN to see how well it performs against DE-CNN. We used aspect embedding instead of domain embedding in the input layer and replaced the final CRF layer with MaxPooling Layer. Results are reported using author's code\footnote{https://github.com/howardhsu/DE-CNN} \cite{xu2018double}. 
\subsubsection{AF-LSTM} 
AF-LSTM incorporates aspect information for learning attention on the sentence using associative relationships between words and aspect \cite{tay2018learning}. 
\subsubsection{GCAE} 
GCAE adopts gated convolution layer for learning aspect representation which is integrated into sentence representation through another gated convolution layer. This model reported results for four sentiment labels. We ran the experiment using author's code\footnote{https://github.com/wxue004cs/GCAE} and reported results for three sentiment labels \cite{xue2018aspect}. 
\subsection{Implementation}
Every word in the input sentence is converted to a 300 dimensional vector using pretrained word embeddings. The dimension of positional embedding is set to 25 which is initialized randomly and updated during training. The hidden units of LSTM are set to 100. The number of hidden units in the layer fully connected to LSTM is 50 and the layer fully connected to positional embedding layer is 50. The number of hidden units in the penultimate fully connected layer is set to 64. We apply a dropout \cite{srivastava2014dropout} with a probability 0.5 on this layer. A batch size 20 is considered and the model is trained for 30 epochs. Adam \cite{kingma2014adam} is used as the optimizer with an initial learning rate 0.001.

\section{Results and Discussion}
The Results are presented in Table \ref{tab:acc}. The Baselines Majority, NBOW and LSTM do not use aspect information for the task at all. Proposed models significantly outperform them. 
\subsection{The Role of Aspect Position}
The proposed model outperforms other recent and popular architectures as well, these architectures use a separate architecture which takes the aspect input distinctly from the sentence input. In doing so they loose the positional information of the aspect within the sentence. We hypothesize that this information is valuable for ATSA and our results reflect the same. Additionally since proposed architecture does not take any additional aspect inputs apart from position, we get a fairer comparison on the benefits of providing aspect positional information over the aspect words themselves.
\subsection{The Role of Decay Functions}
Furthermore, while avoiding learning separate architectures for weightages, decay functions act as good approximates. These functions rely on constants alone and lack any parameters thereby expressing their efficiency. The reason these functions work is because they consider an assumption intrinsic to the nature of most natural languages. It is that description words or aspect modifier words come close to the aspect or the entity they describe. For example in Figure \ref{fig:architecture}, we see the sentence from the Restaurant dataset, ``granted the space is smaller than most, it is the best service you can...''.The proposed model is able to handle this example which has distinct sentiments for the aspects ``space'' and ``service'' due to their proximity from ``smaller'' and ``best'' respectively.  
\section{Conclusion}
In this paper, we propose a novel model for Aspect Based Sentiment Analysis relying on relative positions on words with respect to aspect terms. This relative position information is realized in the proposed model through parameter-less decay functions. These decay functions weight words according to their distance from aspect terms by only relying on constants proving their effectiveness. Furthermore, our results and comparisons with other recent architectures, which do not use positional information of aspect terms demonstrate the strength of the decay idea in proposed model.
\bibliographystyle{splncs04}
\bibliography{nldb}

\begin{thebibliography}{10}
\providecommand{\url}[1]{\texttt{#1}}
\providecommand{\urlprefix}{URL }
\providecommand{\doi}[1]{https://doi.org/#1}

\bibitem{bahdanau2014neural}
Bahdanau, D., Cho, K., Bengio, Y.: Neural machine translation by jointly
  learning to align and translate. arXiv preprint arXiv:1409.0473  (2014)

\bibitem{hu2004mining}
Hu, M., Liu, B.: Mining opinion features in customer reviews. In: AAAI. vol.~4,
  pp. 755--760 (2004)

\bibitem{kingma2014adam}
Kingma, D.P., Ba, J.: Adam: A method for stochastic optimization. arXiv
  preprint arXiv:1412.6980  (2014)

\bibitem{kiritchenko2014nrc}
Kiritchenko, S., Zhu, X., Cherry, C., Mohammad, S.: Nrc-canada-2014: Detecting
  aspects and sentiment in customer reviews. In: Proceedings of the 8th
  International Workshop on Semantic Evaluation (SemEval 2014). pp. 437--442
  (2014)

\bibitem{klambauer2017self}
Klambauer, G., Unterthiner, T., Mayr, A., Hochreiter, S.: Self-normalizing
  neural networks. In: Advances in neural information processing systems. pp.
  971--980 (2017)

\bibitem{nguyen-shirai-2015-phrasernn}
Nguyen, T.H., Shirai, K.: {P}hrase{RNN}: Phrase recursive neural network for
  aspect-based sentiment analysis. In: Proceedings of the 2015 Conference on
  Empirical Methods in Natural Language Processing. pp. 2509--2514. Association
  for Computational Linguistics, Lisbon, Portugal (Sep 2015).
  \doi{10.18653/v1/D15-1298}, \url{https://www.aclweb.org/anthology/D15-1298}

\bibitem{pang2002thumbs}
Pang, B., Lee, L., Vaithyanathan, S.: Thumbs up?: sentiment classification
  using machine learning techniques. In: Proceedings of the ACL-02 conference
  on Empirical methods in natural language processing-Volume 10. pp. 79--86.
  Association for Computational Linguistics (2002)

\bibitem{pontiki-etal-2014-semeval}
Pontiki, M., Galanis, D., Pavlopoulos, J., Papageorgiou, H., Androutsopoulos,
  I., Manandhar, S.: {S}em{E}val-2014 task 4: Aspect based sentiment analysis.
  In: Proceedings of the 8th International Workshop on Semantic Evaluation
  ({S}em{E}val 2014). pp. 27--35. Association for Computational Linguistics,
  Dublin, Ireland (Aug 2014). \doi{10.3115/v1/S14-2004},
  \url{https://www.aclweb.org/anthology/S14-2004}

\bibitem{schmitt-etal-2018-joint}
Schmitt, M., Steinheber, S., Schreiber, K., Roth, B.: Joint aspect and polarity
  classification for aspect-based sentiment analysis with end-to-end neural
  networks. In: Proceedings of the 2018 Conference on Empirical Methods in
  Natural Language Processing. pp. 1109--1114. Association for Computational
  Linguistics, Brussels, Belgium (Oct-Nov 2018),
  \url{https://www.aclweb.org/anthology/D18-1139}

\bibitem{socher-etal-2013-recursive}
Socher, R., Perelygin, A., Wu, J., Chuang, J., Manning, C.D., Ng, A., Potts,
  C.: Recursive deep models for semantic compositionality over a sentiment
  treebank. In: Proceedings of the 2013 Conference on Empirical Methods in
  Natural Language Processing. pp. 1631--1642. Association for Computational
  Linguistics, Seattle, Washington, USA (Oct 2013),
  \url{https://www.aclweb.org/anthology/D13-1170}

\bibitem{srivastava2014dropout}
Srivastava, N., Hinton, G., Krizhevsky, A., Sutskever, I., Salakhutdinov, R.:
  Dropout: a simple way to prevent neural networks from overfitting. The
  Journal of Machine Learning Research  \textbf{15}(1),  1929--1958 (2014)

\bibitem{tang2015effective}
Tang, D., Qin, B., Feng, X., Liu, T.: Effective lstms for target-dependent
  sentiment classification. arXiv preprint arXiv:1512.01100  (2015)

\bibitem{tay2018learning}
Tay, Y., Tuan, L.A., Hui, S.C.: Learning to attend via word-aspect associative
  fusion for aspect-based sentiment analysis. In: Thirty-Second AAAI Conference
  on Artificial Intelligence (2018)

\bibitem{AAAI1816570}
Tay, Y., Tuan, L.A., Hui, S.C.: Learning to attend via word-aspect associative
  fusion for aspect-based sentiment analysis (2018)

\bibitem{wang2016attention}
Wang, Y., Huang, M., Zhao, L., et~al.: Attention-based lstm for aspect-level
  sentiment classification. In: Proceedings of the 2016 conference on empirical
  methods in natural language processing. pp. 606--615 (2016)

\bibitem{xu2018double}
Xu, H., Liu, B., Shu, L., Yu, P.S.: Double embeddings and cnn-based sequence
  labeling for aspect extraction. arXiv preprint arXiv:1805.04601  (2018)

\bibitem{xue2018aspect}
Xue, W., Li, T.: Aspect based sentiment analysis with gated convolutional
  networks. arXiv preprint arXiv:1805.07043  (2018)

\end{thebibliography}
\end{document}